\documentclass[sigconf]{acmart}
\AtBeginDocument{%
  }

\allowdisplaybreaks % allow page breaks within equations

\copyrightyear{2023} 
\acmYear{2023} 
\setcopyright{acmlicensed}
\acmConference[MM '23]{Proceedings of the 31st ACM International Conference on Multimedia}{October 29-November 3, 2023}{Ottawa, ON, Canada}
\acmBooktitle{Proceedings of the 31st ACM International Conference on Multimedia (MM '23), October 29-November 3, 2023, Ottawa, ON, Canada}
\acmPrice{15.00}
\acmDOI{10.1145/3581783.3611805}  

\acmISBN{979-8-4007-0108-5/23/10}

\usepackage{balance}

\usepackage{multirow}

\usepackage{enumitem}

\usepackage{graphicx}  %插入图片的宏包
\usepackage{float}  %设置图片浮动位置的宏包
\usepackage{subfigure}  %插入多图时用子图显示的宏包
\usepackage{amsmath}

\begin{document}

\title[Multimodal Prompt Transformer with Hybrid Contrastive Learning]{Multimodal Prompt Transformer with Hybrid Contrastive Learning for Emotion Recognition in Conversation}

\author{Shihao Zou}
\orcid{0000-0002-6648-3995}
\affiliation{%
  \institution{Chongqing University of Technology}
  \streetaddress{College of Computer Science and Engineering}
  \city{Chongqing}
  \country{China}
}
\email{z_sh9904@163.com}

\author{Xianying Huang}
% \orcid{0000-0002-3667-6198}
\authornote{Corresponding author}
\affiliation{%
  \institution{Chongqing University of Technology}
  \streetaddress{College of Computer Science and Engineering}
  \city{Chongqing}
  \country{China}}
\email{wldsj_cqut@163.com}

\author{Xudong Shen}
% \orcid{0009-0008-0583-9717}
\affiliation{%
  \institution{Chongqing University of Technology}
  \streetaddress{College of Computer Science and Engineering}
  \city{Chongqing}
  \country{China}
}
\email{sxd_cqut@163.com}

%%
%% By default, the full list of authors will be used in the page
%% headers. Often, this list is too long, and will overlap
%% other information printed in the page headers. This command allows
%% the author to define a more concise list
%% of authors' names for this purpose.
\renewcommand{\shortauthors}{Shihao Zou, Xianying Huang, \& Xudong Shen}
%% No italics
%% The abstract is a short summary of the work to be presented in the
%% article.
\begin{abstract}
Emotion Recognition in Conversation (ERC) plays an important role in driving the development of human-machine interaction. Emotions can exist in multiple modalities, and multimodal ERC mainly faces two problems: (1) the noise problem in the cross-modal information fusion process, and (2) the prediction problem of less sample emotion labels that are semantically similar but different categories. To address these issues and fully utilize the features of each modality, we adopted the following strategies: first, deep emotion cues extraction was performed on modalities with strong representation ability, and feature filters were designed as multimodal prompt information for modalities with weak representation ability. Then, we designed a Multimodal Prompt Transformer (MPT) to perform cross-modal information fusion. MPT embeds multimodal fusion information into each attention layer of the Transformer, allowing prompt information to participate in encoding textual features and being fused with multi-level textual information to obtain better multimodal fusion features. Finally, we used the Hybrid Contrastive Learning (HCL) strategy to optimize the model's ability to handle labels with few samples. This strategy uses unsupervised contrastive learning to improve the representation ability of multimodal fusion and supervised contrastive learning to mine the information of labels with few samples. Experimental results show that our proposed model outperforms state-of-the-art models in ERC on two benchmark datasets.
\end{abstract}

%%
%% The code below is generated by the tool at http://dl.acm.org/ccs.cfm.
%% Please copy and paste the code instead of the example below.
%%


% \begin{CCSXML}
% <ccs2012>
%    <concept>
%        <concept_id>10002951.10003227.10003251</concept_id>
%        <concept_desc>Information systems~Multimedia information systems</concept_desc>
%        <concept_significance>500</concept_significance>
%        </concept>
%    <concept>
%        <concept_id>10002951.10003227</concept_id>
%        <concept_desc>Information systems~Information systems applications</concept_desc>
%        <concept_significance>300</concept_significance>
%        </concept>
%  </ccs2012>
% \end{CCSXML}

% \ccsdesc[500]{Information systems~Multimedia information systems}
% \ccsdesc[300]{Information systems~Information systems applications}

% \begin{CCSXML}
% <ccs2012>
%    <concept>
%        <concept_id>10010147.10010178</concept_id>
%        <concept_desc>Computing methodologies~Artificial intelligence</concept_desc>
%        <concept_significance>500</concept_significance>
%        </concept>
%    <concept>
%        <concept_id>10010147.10010178.10010179</concept_id>
%        <concept_desc>Computing methodologies~Natural language processing</concept_desc>
%        <concept_significance>500</concept_significance>
%        </concept>
%  </ccs2012>
% \end{CCSXML}

% \ccsdesc[500]{Computing methodologies~Artificial intelligence}
% \ccsdesc[500]{Computing methodologies~Natural language processing}

%%
%% Keywords. The author(s) should pick words that accurately describe
%% the work being presented. Separate the keywords with commas.
%% A "teaser" image appears between the author and affiliation
%% information and the body of the document, and typically spans the
%% page.

% \received{20 February 2007}
% \received[revised]{12 March 2009}
% \received[accepted]{5 June 2009}

%%
%% This command processes the author and affiliation and title
%% information and builds the first part of the formatted document.
\maketitle

\section{Introduction}
With the rapid development of social networks, there has been a lot of attention given to building dialogue systems that can understand user emotions and intentions and engage in effective dialogue interaction. Emotion Recognition in Conversation (ERC) is a task that assigns emotional labels with contextual relationships to each utterance made by speakers during a conversation. As a relevant task for dialogue systems, ERC has made important contributions to the development of engaging, interactive, and empathetic dialogue systems by analyzing user emotions in the context of a conversation.  It has greatly propelled the advancement of human-machine interaction \cite{MaNXC20}.

\begin{figure}[h]
  \centering
  \includegraphics[width=\linewidth]{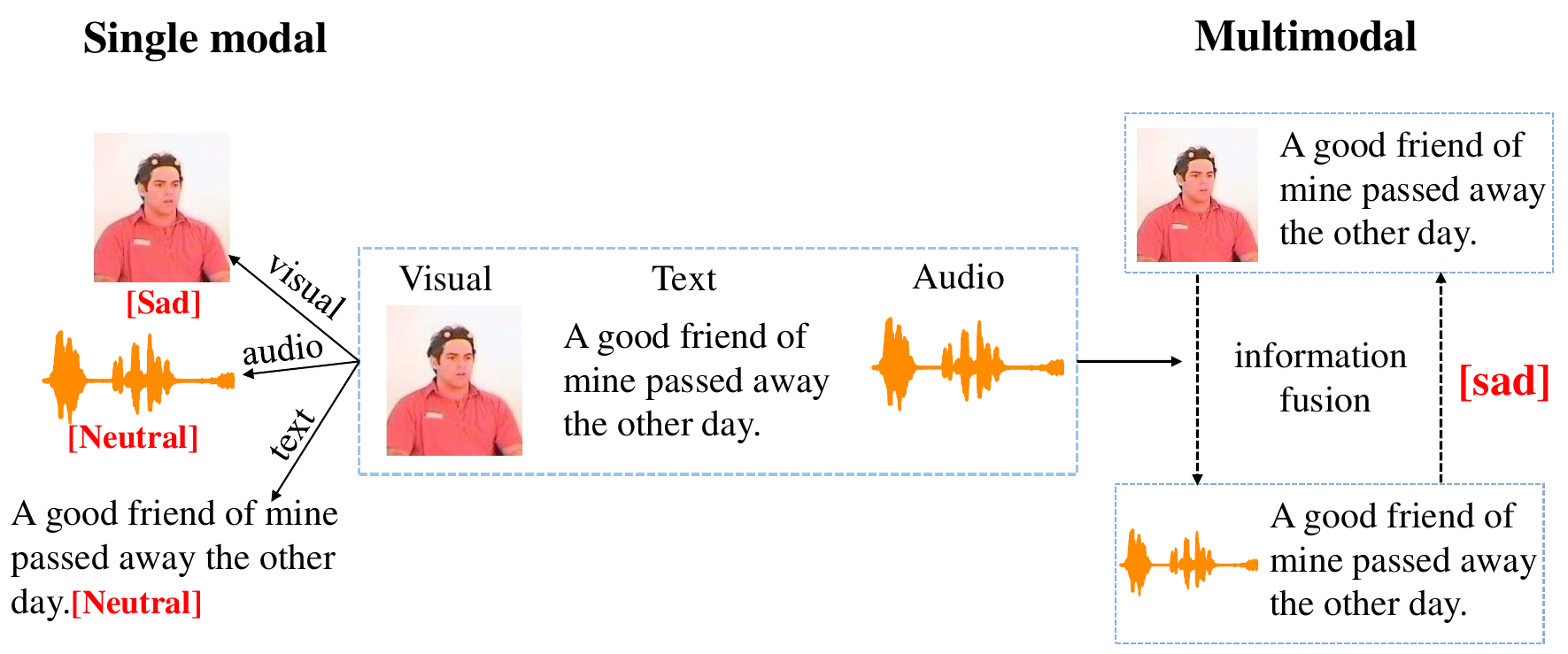}
  \caption{Example of unimodal vs. multimodal in IEMOCAP.}
  \label{fig:1}
\end{figure}

In the development of ERC, early ERC studies \cite{dialoguernn/MajumderPHMGC19,dialoguegcn/GhosalMPCG19,rgat/IshiwatariYMG20} focused mainly on research methods that only used text features. However, researchers found that text contains many types of utterances that are difficult for models to fully understand, such as irony, and that emotions in conversation also change with the dynamic interactive process. Since it is difficult to fully understand the conversational context, using only text is insufficient to capture emotional cues in the conversation for accurate emotion prediction. Therefore, people have attempted to expand the model's input from a single modality to multiple modalities (such as visual, audio, videos, etc.). As shown in Figure \ref{fig:1}, using a single modality always leads to prediction errors, but using more types of modality information, the problem of semantic limitations of using only text can be mitigated. However, after introducing multiple modalities, since each modality is in a different semantic space, cross-modal information interactions will introduce a lot of noise if the semantic gaps in them are not considered. For example, Hu et al. \cite{mmgcn/HuLZJ20} constructed a modality interaction graph by treating different modality features as nodes. Although this method uses multiple modality information, it does not consider the semantic gap between modalities and directly performs information interaction across modalities, leading to suboptimal fusion effects and affecting the final emotion prediction. 

In addition, because the ERC datasets contained many semantically similar but few sample labels (e.g., fear and disgust, happy and excited), some previous studies \cite{mmdfn/HuHWJM22} concluded that the number of these label samples was less and their results were not statistically significant, so they were merged into other similar emotion category samples. Although this approach can improve the accuracy of emotion prediction, from a psychological perspective \cite{emotion1,emotion2,emotion3} , each emotion category reflects its independent emotional intensity, and simply combining emotion categories leads to oversimplification and may not accurately capture the emotional complexity and diversity of experience. Furthermore, since each emotion label represents the user's emotional state, accurate predictions should be obtained for each category label in real-world scenarios.

To address the above problems, we propose a new model for ERC, namely Multimodal Prompt Transformer with Hybrid Contrastive Learning (MPT-HCL). Firstly, in order to extract emotional cues from the text, we construct exclusive relationship graphs from both the speaker and context levels to extract emotional cues at different levels. For the audio and visual modalities, we filter the features using a designed modal feature filter, which filters out low-level features with more noise and retains high-level features with valid information. Due to previous research demonstrating the importance of text modality \cite{kbs/ZouHSL22,mtag/YangWYZRZPM21}, we consider text as the main modality feature and refer to the filtered features of audio and visual as textual prompt information. Then, we use MPT for information interaction between modalities. MPT embeds the interaction of multiple modalities into each attention layer of the Transformer, allowing the multimodal prompt information to participate in text feature encoding and fusion with multi-level text information, thus reducing noise generation and using the multimodal fusion result for emotion prediction of each utterance. Finally, to better optimize the model's performance, on one hand, we use unsupervised contrastive learning (UCL) to repeatedly extract mutual information \cite{mim/BelghaziBROBHC18} between the fusion features and each unimodal modality, in order to mine the relationship between modalities and optimize the fusion feature representation. On the other hand, we use supervised contrastive learning (SCL) to mine the relationship between fused features and labels in the sample. As SCL aggregates less sample labels by treating all samples with the same label in a batch as positive examples, it enhances the presence of less sample labels in the batch. Through this hybrid contrastive learning method, the feature representation of multimodal fusion can be optimized, thus effectively improving the accuracy of prediction for less sample labels. In summary, the main contributions of this paper are summarized as follows:
\begin{itemize}[itemsep=0pt,topsep=0pt,parsep=0pt,leftmargin=*]
    \item We propose a novel approach of using filtered modality information as multimodal prompt information, and designing a multimodal prompt transformer for cross-modal information interaction to enhance the fusion effect of multiple modalities.   
    \item For the first time in multimodal ERC, we introduced hybrid contrastive learning to separately explore the information between the fused modal features and each modality, as well as the information in the labels of the samples.
    \item We propose a new ERC model, MPT-HCL, which adopts a multimodal fusion method with hybrid contrastive learning to improve context understanding and the accuracy of multimodal ERC.
    \item We conducted extensive experiments on two public benchmark multimodal datasets, including IEMOCAP \cite{Iemocap} and MELD \cite{Meld}. The results showed that our proposed MPT-HCL model is more effective and superior to all SOTA baseline models.
\end{itemize}

\section{Related WORKS}
\subsection{Emotion Recognition in Conversation}
Emotion recognition in conversation, as an important research area in natural language processing (NLP), has received extensive attention in recent years. Existing research on ERC mainly has two types of data input, text-based and multimodal-based: (1) Text-based: DialogueGCN \cite{dialoguegcn/GhosalMPCG19} uses graph networks for modeling dependencies between self- and inter- of speakers, which effectively solves the DialogueRNN \cite{dialoguernn/MajumderPHMGC19} suffers from the context propagation problem; Ishiwatari et al. \cite{rgat/IshiwatariYMG20} proposes that R-GAT with relational location encoding not only captures the dependency relations between speakers, but also provides sequential information about the relational graph structure; Shen et al. \cite{dag/ShenWYQ20} designs a directed acyclic graph (DAG) neural network to encode the utterance to better model the intrinsic structure in the conversation and thus explicitly model the information of each speaker in the conversation; TODKAT \cite{todkat/ZhuP0ZH20} utilizes the encoder-decoder architecture, which combines the representation of topic information with common-sense information in ERC; (2) Multimodal-based: ICON \cite{icon/HazarikaPMCZ18}  and CMN \cite{cmn/HazarikaPZCMZ18} both model information in conversation sequences by GRU; MulT \cite{Mult/TsaiBLKMS19} uses Transformer's \cite{transformer} fusion approach of the basic module-multihead attention mechanism to achieve cross-modal information fusion by using different modalities as query, key, and value in attention respectively; Li et al. \cite{emocaps/LiTZZ22} proposes a new structure called Emoformer to extract multimodal emotion vectors from different modalities and fuse them with sentence vectors into an emotion capsule; MM-DFN \cite{mmdfn/HuHWJM22} uses a new multimodal dynamic fusion network to capture dynamic changes of contextual information in different semantic spaces.

% HCL-ERC \cite{hcl-erc/YangSMC22} introduces curriculum learning into the field of ERC for the first time by setting up two levels of curriculums to divide the data. Based on some of the models mentioned above,the performance has been greatly improved in both.

\subsection{Contrastive Learning}
In the field of computer vision, SimCLR \cite{simclr/ChenK0H20} optimizes contrast loss by using images obtained from the same image by randomly different data enhancement as positive samples and other images as negative samples. In natural language pre-training, ConSERT \cite{consert/YanLWZWX20} introduces self-supervised contrast loss in the fine-tuning phase of BERT \cite{bert/DevlinCLT19} in order to address the poor performance of sentence representation in semantic similarity tasks; Li et al. \cite{cogbart/LiYQ22} uses supervised contrast learning on top of BART \cite{bart/LewisLGGMLSZ20} as the backbone network to make different emotions to be mutually exclusive to better identify similar emotions. In terms of multimodal learning, TupleInfoNCE \cite{tupleinfo/LiuFZ0FY21} is a method for learning representations of multimodal data using contrast loss that learns complementary synergies between modalities; MMIM \cite{mmim/HanCP21} maintains task-relevant information by maximizing mutual information in single-peaked input pairs.

\section{Task Definition}
In ERC, the data consists of multiple conversations $\{c_1,c_2,\ldots,c_N\}$, and each conversation consists of a series of utterances $c_i=[u_1,u_2,\\ \ldots,u_m]$, where $N$ is the number of conversations in a batch of data, and $m$ is the number of utterances in the i-th conversation. Each utterance $u_i$ consists of $n_i$ tokens, i.e., $\{w_{i1},w_{i2},\ldots,w_{in_i}\}$ . Each conversation has $M$ speakers $P=\{p_1,p_2,\ldots,p_M\},(M \ge 2)$ and each utterance is spoken by a speaker $p_{s(u_i)}$, where the function $s(\cdot)$ maps the index of the utterance to the corresponding speaker. The discrete value $y_i \in S$ is used to represent the emotion labels of $u_i$, where $S$ is the set of emotion labels. The purpose of ERC is to input a conversation and identify the correct emotion classification for each utterance in the conversation from the set of emotion labels. For each utterance $u_i$ , we extract the multimodal features $u_i=\{u_i^m\},m \in \{a,v,t\} $ . Here, $u_i^a \in \mathbb{R}^{d_a}$,$u_i^v \in \mathbb{R}^{d_v}$ and $u_i^t \in \mathbb{R}^{d_t}$ are the audio, visual and text feature representations of the utterance, respectively. And $\{d_m\}, m \in \{a,v,t\}$ is the feature dimension of each modality.

\begin{figure*}[h]
  \centering
  \includegraphics[width=\linewidth]{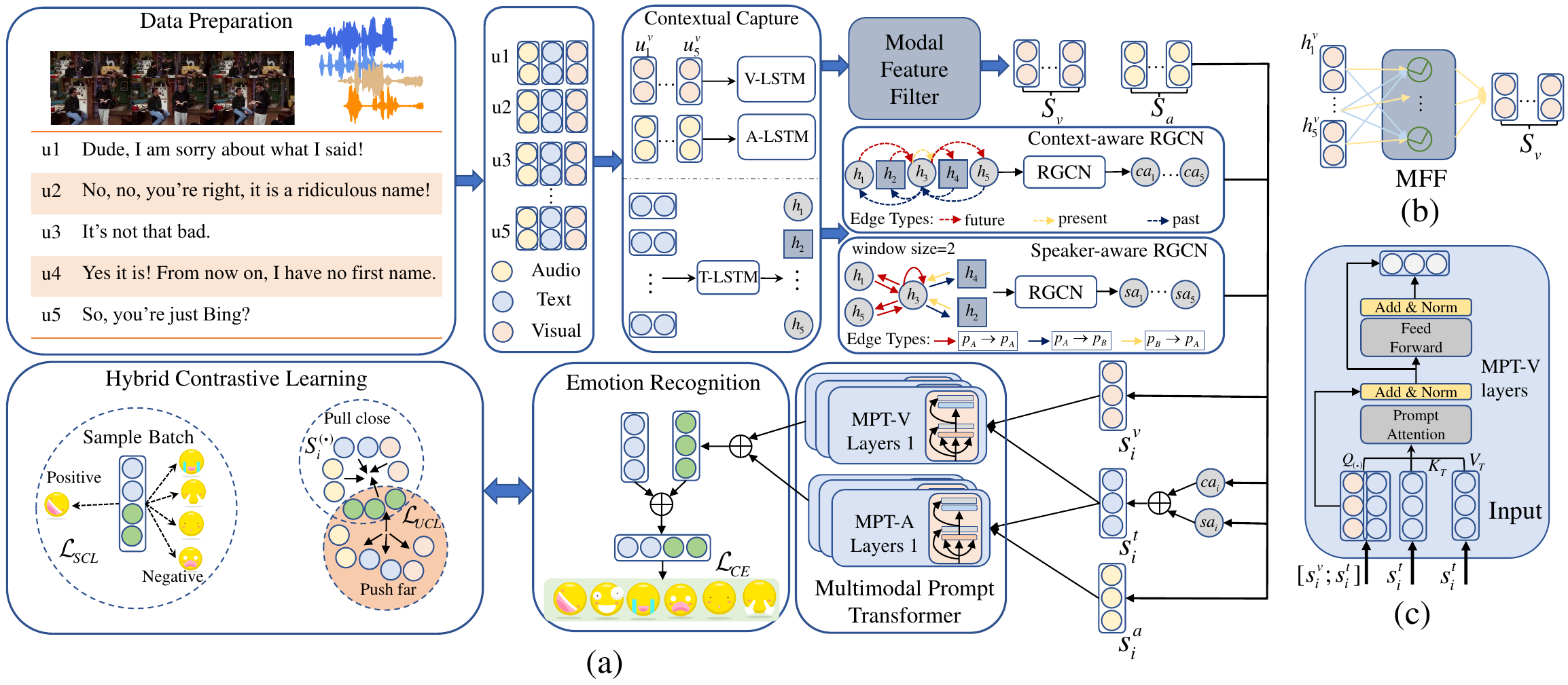}
  \caption{(a) shows the specific structure of MPT-HCL. (b) illustrates the architecture of the Modality Feature Filter (MFF) module, showcasing its input and output representations. (c) provides a detailed view of MPT-V, designed specifically for the visual modality.}
  \label{fig:2}
\end{figure*}

\section{Multimodal Prompt Transformer with Hybrid Contrastive Learning}
In this section, we first introduce the learning methods for different modalities. Then, we present the designed cross-modal fusion method, the Multimodal Prompt Transformer (MPT). Finally, we show our hybrid contrastive learning (HCL) optimization strategy. The architecture of our model MPT-HCL is shown in Figure~\ref{fig:2}.
\subsection{Contextual Capture}
For each modality, we use Bi-LSTM for contextual capture, which is calculated as follows:
\begin{equation}
    \begin{aligned}
        &h_i^m = BiLSTM^m(u_i^m,h_{i-1}^m)\\
        &H_m = \{h_i^m\}_{i=1}^L \in \mathbb{R}^{L \times d_m},m\in \{a,v,t\}
    \end{aligned}
\end{equation}
where $h_{i}^m$ is the i-th hidden layer state of Bi-LSTM. $L$ denotes the conversation sequence length in the batch, and we use $H_t=\{h_i^t\}_{i=1}^L$ as the initial representation of the node.

\subsection{Modal Feature Filter}
To reduce noise in the modal features, we have designed a modal feature filter (MFF) that selects the features with more useful information while filtering out the low-level features with excessive noise. This allows us to obtain the multimodal prompt information. The feature filter is mainly composed of a dynamic gating module. The dynamic gate predicts a normalized vector for each modality, which represents the degree to which information needs to be obtained from that modality. We take the visual modality as an example, in the dynamic gate, $z_v^{(l)}$ represents the vector that indicates the degree to which the visual modality provides information to the text modality in the l-th layer of the Transformer. We first calculate the logit of the gating signal $\theta_v$ :
\begin{equation}
    \theta_v = f(W_l(\frac{1}{L}\sum_{i=1}^{L} P(h_i^v)))
\end{equation}
where $f(\cdot)$ denotes the Leaky\_ReLU activation function, $P(\cdot)$ represents the average pooling, which is used to generate the average vector by weighting the average of the utterances in the current batch, and $W_l$ is the parameter matrix of the linear layer. After that, a probability vector $z_v$ is generated for the visual feature representation as follows:
\begin{equation}
    z_v = softmax(\theta_v)
\end{equation}

Based on the probability results obtained from the dynamic gating, we obtain the final aggregated visual gating feature $V_{gate}$, and then feed the aggregation result into the lower and upper projection layers for better information fusion:
\begin{equation}
    \begin{aligned}
        &V_{gate} = z_vH_v \\
        &V_d = \sigma(W_dV_{gate}+b_d) \\
        &S_v = W_uV_d + b_u
    \end{aligned}
\end{equation}
where $S_v\in \mathbb{R}^{L \times d}$ is the visual prompt feature representation obtained by the modal feature filter, the audio prompt information $S_a\in \mathbb{R}^{L \times d}$ is obtained in the same way as the visual, $d$ is the dimension of the modal features after alignment. $\sigma(\cdot)$ is the sigmoid activation function, and $\{W_d,W_u,b_d,b_u\}$ are the trainable parameters.

\subsection{Speaker \& Context-aware RGCN }
Inspired by \cite{GCNet}, but we don't use it for the same purpose. We design speaker-aware RGCN (Sa-RGCN) and context-aware RGCN (Ca-RGCN) as the core modules for emotion cue extraction in text modality. In these modules, edges are used to measure the importance of connections between nodes. The type of edge defines the method of propagation of different information between nodes. Sa-RGCN and Ca-RGCN have the same edges, but each edge represents a different dependency, and we will describe the composition of these two modules in detail next.

% We design speaker-aware RGCN (Sa-RGCN) and context-aware RGCN (Ca-RGCN) as the core modules that effectively and sufficiently model the dynamic changes of emotional cues in the conversation.

\textbf{Edge:} For each node, its interaction with the context nodes should be considered. If each node $v_i$ interacts with all the context nodes $v_j$, the constructed graph contains a large number of edges, which usually leads to the problem of computational difficulties due to the huge number of parameters. To solve this problem, we fix the context window size to $w$ , so that each node $v_i$ interacts with only $\{v_j\}_{j=max(i-w,1)}^{min(i+w,L)}$ context nodes. In our model implementation, we perform the selection of $w$ within $w \in \{1,2,3,4\}$ and the edges $e_{ij}$ denote nodes $v_i$ to $v_j$.

\textbf{Speaker-aware RGCN:} Sa-RGCN uses different speakers and their spoken utterances to capture the dependencies of the speakers in a conversation. Specifically, we assign a speaker identifier $\alpha_{ij} \in$ \text{ \LARGE $\alpha$} to each edge $e_{ij}$. Here \text{ \LARGE $\alpha$} denotes the set of speaker types in the conversation, and \text{ \LARGE $|\alpha|$} denotes the number of \text{ \LARGE $\alpha$}. For each edge $e_{ij}$ , $\alpha_{ij}$ serves as the set of $p_{s(u_i)} \rightarrow p_{s(u_j)}$, where $p_{s(u_i)}$ and $p_{s(u_j)}$ denote the speaker identifiers of $u_i$ and $u_j$ , respectively.

\textbf{Context-aware RGCN:} Ca-RGCN uses contextual information to capture the contextual dependencies in a conversation. Specifically, we assign a context type identifier $\beta_{ij} \in$ \text{ \LARGE $|\beta|$} to each edge $e_{ij}$, where \text{ \LARGE $\beta$} denotes the set of context types in the conversation. Based on the relative positions of $u_i$ and $u_j$ in the conversation, we will perform $\beta_{ij}$ value determination from $\{past,present,future\}$, so \text{ \LARGE $|\beta|$}=3.

\textbf{Graph learning:} We use RGCN to aggregate the neighbor information in the graph. For Sa-RGCN and Ca-RGCN, we pass different information through the edges, and the parameters depend on the type of the edges. The calculation is shown as follows:
\begin{equation}
    \begin{aligned}
        &sa_i = ReLU(\sum_{r\in \text{\LARGE $\alpha$}}\sum_{j \in N_i^r} \frac{1}{|N_i^r|}W_r^sh_j^t) \\
        &ca_i = ReLU(\sum_{r \in \text{\LARGE $\beta$}}\sum_{j \in N_i^r} \frac{1}{|N_i^r|}W_r^ch_j^t)
    \end{aligned}
\end{equation}
where $sa_i$ and $ca_i$ denote the outputs of nodes in Sa-RGCN and Ca-RGCN, respectively. $N_i^r$ denotes the set of all neighbor nodes of $v_i$ under relation $r$, and $|N_i^r|$ is the number of $N_i^r$. $W_r^s$ and $W_r^c$ denote the parameter matrices of Sa-RGCN and Ca-RGCN at relation $r$, respectively.

After obtaining the speaker and context dependencies in the text, we fuse them to obtain the enhanced textual feature representation $s_i^t$ , which is computed as follows:
% \vspace{-\baselineskip} % reduce space by one line
\begin{equation}
    s_i^t = sa_i + ca_i
\end{equation}

\subsection{Multimodal Interaction}
We consider the audio and visual features obtained by the modal feature filters module as textual prompt information, and update all text states with the prompt information when executing cross-modal attention sequentially. In this way, the final fusion feature will have the ability to encode both the context and cross-modal semantic information, which effectively alleviates the problem of noise generated by unrelated elements in the process of multimodal fusion.

\textbf{Prompt Attention.} First, we project the textual feature sequence $S_t=\{s_i^t\}_{i=1}^L \in \mathbb{R}^{L \times d}$ enhanced with emotional cues into the query/key/value vector, i.e., $Q=S_tW_Q, K=S_tW_K, V=S_tW_V$, where $W_{(\cdot)}$ is the trainable parameter matrix. Then we prepend the visual prompt feature sequences $S_v=\{s_i^v\}_{i=1}^L \in \mathbb{R}^{L \times d}$ and audio prompt feature sequences $S_a=\{s_i^a\}_{i=1}^L \in \mathbb{R}^{L \times d}$ obtained from the modal feature filters to each attention layer of the Transformer, respectively, and we call this proposed prepending method Prompt Attention (PA), by which the prompt features are used for effective cross-modal interaction, as implemented in the following equation shown:
\begin{equation}
    \begin{aligned}
        Y_{tv} &= PA([S_v;S_t],S_t,S_t) \\
        &=softmax(\frac{[S_v;Q]K^T}{\sqrt{d}})V
    \end{aligned}
\end{equation}
where $[;]$denote feature concatenate, $Y_{tv} \in \mathbb{R}^{L \times d}$ is the result of the weight of a layer of PA. In this approach, multiple attentions are combined to obtain the output results of the multihead attention layer as follows:
\begin{equation}
    Multihead(Y_{tv})=Concat(Y_{tv}^1,\ldots,Y_{tv}^n)W
    \label{con:eq1}
\end{equation}
where $Y_{tv}^1,\ldots,Y_{tv}^n$ is the output of each attention layer, $n$ is the number of attention layers, and $W$ is the trainable parameter matrix. PA is used to perform cross-modal interaction processes between text modalities and multimodal prompt information.

\textbf{Multimodal Prompt Transformer.} Based on the above PA, we design the Multimodal Prompt Transformer (MPT) with the structure shown in Figure~\ref{fig:2}(c). Since we embed multimodal interactions into each attention layer of the Transformer, visual and audio features can participate in the encoding of text features and fuse with multi-layered text information. Low-level syntactic features encoded by the shallow Transformer layer and high-level semantic features encoded by the deep Transformer layer \cite{attention/PetersNIGCLZ18,transformer} interact with the visual and audio prompt features, enabling the fusion of information across modalities. In addition depending on the multimodal prompt features used, as more effective feature representation can fully exploit the fusion effect between modalities.

We use a residual connectivity layer with regularization to normalize the output of the multihead attention layer of Eq.\eqref{con:eq1} and use a position feedforward sublayer to obtain the output of the attention:
\begin{equation}
    \begin{aligned}
        &N = Norm(Y_{tv}+Multihead(Y_{tv})) \\
        &F = max(0,NW_1+b_1)W_2+b_2 \\
        &G = Norm(F+Multihead(F))
        \label{con:eq2}
    \end{aligned}
\end{equation}
where $W_1,W_2$ is the weight parameter and $b_1,b_2$ is the bias parameter. In this process each modal prompt feature is continuously updated with the feature of the text by the above method to obtain better fusion results, and finally we use self-attention to collect the sequence information of the modal fusion features to obtain the multimodal fusion result $X \in \mathbb{R}^{L \times d}$ for the current conversation:
\begin{equation}
    X = Attention(G)
    \label{con:eq3}
\end{equation}

Connecting the above Eq.\eqref{con:eq1}  to \eqref{con:eq3}, we can input different modal prompt features and text features to MPT to obtain the modal fusion feature representation:
\begin{equation}
    \begin{aligned}
        &X_{tv} = MPT([S_v;S_t],S_t,S_t) \\
        &X_{ta} = MPT([S_a;S_t],S_t,S_t)
    \end{aligned}
\end{equation}

where $S_t,S_v,S_a$ denote the text, visual and audio features obtained by different learning methods as the input to the MPT, and $X_{tv},X_{ta}$ denote the multimodal fusion results obtained after prompting the text with visual prompt features and audio prompt features, respectively. We concatenate the outputs from different MPTs to obtain the fused features $X_{mpt} \in \mathbb{R}^{L \times 2d}$ :
\begin{equation}
    X_{mpt} = X_{tv} \oplus X_{ta}
\end{equation}

Finally, in order to fully utilize the text feature representation, we combine $X_{mpt}$ with the text modality to obtain the final representation of the current utterance used for emotion prediction $X_{fusion}$:
\begin{equation}
    X_{fusion} = S_t \oplus X_{mpt}
\end{equation}

 \subsection{Hybrid Contrastive Learning}
 \textbf{Unsupervised contrastive learning.} Although a better representation of the fusion features is obtained by the MPT, the relationship between each unimodal modal and the fusion features is not entirely explored, so we use unsupervised contrastive learning(UCL) to exploit the connection between them and thus optimize the obtained fusion features. We repeat the mutual information maximization between the fusion results and the input modalities, and the optimization goal is to fuse the network from each unimodal modal to the fusion features. Since we now obtain the multimodal fusion result $X_{mpt}$ by the constructed MPT network , but the mining of the connection from the fusion feature $X_{mpt}$ to each unimodal input $S_x,x \in \{a,v,t\}$ is missing. So we follow the operation of \cite{oord} and use the score function $Score(\cdot)$ with normalized prediction and true vector to measure the connection between them as follows:
\begin{equation}
    \begin{aligned}
        &Score(s_x,X_{mpt}) = exp(\overline s_x(\overline G_{\varphi}(X_{mpt}))^T)\\
        &\overline G_{\varphi}(X_{mpt}) = \frac{G_{\varphi}(X_{mpt})}{||G_{\varphi}(X_{mpt})||_2}, \overline s_x = \frac{s_x}{||s_x||_2}
    \end{aligned}
\end{equation}
where $G_{\varphi}$ is a neural network with parameter $\varphi$ that generates a prediction for $s_x$ from $X_{mpt}$, and $||\cdot||_2$ is the Euclidean norm. We treat all other representations of the modality in the same batch as negative samples, and thus calculate the loss between the individual modality and the fused features:
\begin{equation}
    \mathcal{L}(X_{mpt},S_x) = -\mathbb{E}_s \Big [ log\frac{Score(X_{mpt},s_x^i)}{\sum_{s_x^j \in S_x}Score(X_{mpt},s_x^j)} \Big ]
\end{equation}

Finally, the loss function of UCL consists of the losses between the fused features $X_{mpt}$ (noted here as $x$) and the text, visual and audio, respectively:
\begin{equation}
    \mathcal{L}_{UCL} = \mathcal{L}^{x,v}+\mathcal{L}^{x,a}+\mathcal{L}^{x,t}
\end{equation}
 \textbf{Supervised contrastive learning.} Supervised contrastive learning(SCL) assumes that attention will be paid to certain key labels, and then treats all samples in the batch with the same label as positive samples and those with different labels as negative samples by making full use of the label information. Specifically for ERC, since the number of samples in each category in the dataset is very imbalanced, such that the information of these samples is masked in the process of calculating the loss. In addition, if only one sample exists in a category batch, it cannot be directly applied to the loss calculation. Therefore, in order to ensure that a sufficient number of feature representations are available for the loss calculation each time, we combine the obtained multimodal fusion features and text features at the position of sequence length. For a batch with $L$ training samples, we can obtain $2L$ samples in this way, after which the SCL loss $\mathcal{L}_{SCL}$ is calculated as follows:
\begin{equation}
    \begin{aligned}
        C=[S_t;X_{mpt}]
    \end{aligned}
\end{equation}
 
 \begin{equation}
     \begin{aligned}
         &\mathcal{L}_{SCL} = \sum_{i \in I}\frac{-1}{|P(i)|}\sum_{p \in P(i)}SIM(p,i) \\
         &SIM(p,i) = log \frac{exp((C_i \cdot C_p)/\tau)}{\sum_{a \in A(i)}exp(C_i \cdot C_p / \tau)}
     \end{aligned}
 \end{equation}
 where $C \in \mathbb{R}^{2L \times d}, i \in I=\{1,2,\ldots, 2L\}$ denotes the index of samples in a batch, $\tau \in R^+$ denotes the temperature coefficient used to control the distance between samples, $P(i)=I_{j=i}-\{i\}$ denotes the samples with the same emotion category as $i$ but excluding $i$ itself, $|P(i)|$ denotes the number of samples, and $A(i)=I-\{i\}$ denotes the samples in a batch other than itself.

\subsection{Model Training}
The loss of the model training process consists of three components: the logarithmic loss of standard cross-entropy $\mathcal{L}_{CE}$ , the supervised contrastive loss $\mathcal{L}_{SCL}$ and the unsupervised contrastive loss $\mathcal{L}_{UCL}$.
\begin{equation}
    \begin{aligned}
        &P_i = softmax(W_sX_{fusion}+b_s) \\
        &\hat{y}_i = argmax(P_i)
    \end{aligned}
\end{equation}
\begin{equation}
    \mathcal{L}_{CE} = -\frac{1}{N}\sum_{i=1}^{N}\sum_{j=1}^{J}y_{i,j} \cdot log  \hat{y}_{i,j}
\end{equation}
\begin{equation}
    \mathcal{L} = \mathcal{L}_{CE} + \lambda_1 \mathcal{L}_{SCL} + \lambda_2 \mathcal{L}_{UCL}
\end{equation}
where $N$ is the number of conversations, $J$ denotes the number of emotion categories, $y_{i,j}$ is the true emotion label of utterance $i$, $\hat{y}_{i,j}$ denotes the probability distribution that the prediction of utterance $i$ is category $j$, and $\lambda_1, \lambda_2$ denote the weights of supervised and unsupervised contrastive loss, respectively. In the training process we use the Adam \cite{adam} optimizer with stochastic gradient descent to train our network model.

\section{Experiment }
\subsection{Datasets and Evaluations}
We evaluated the effectiveness of our model on two benchmark datasets, IEMOCAP \cite{Iemocap} and MELD \cite{Meld}. Both datasets are multimodal ERC datasets containing text, audio, and visual. We divided the datasets according to \cite{mmdfn/HuHWJM22}. The data distribution of two datasets are shown in Table \ref{tab:1} and emotion distribution information of two datasets are shown in Table \ref{tab:2}.

\begin{table}[!ht]
\caption{Data distribution of IEMOCAP and MELD.}
\label{tab:1}
\centering
\renewcommand{\arraystretch}{1.0}
\scalebox{0.95}{
\begin{tabular}{cccccc}
    \toprule[1.2pt]
    \multirow{2}{*}{Dataset} & \multicolumn{2}{c}{Conversation} & \multicolumn{2}{c}{Utterance} & \multirow{2}{*}{Classes}\\
    \cmidrule(r){2-3} \cmidrule(r){4-5} 
    & Train+Val & Test  & Train+Val & Test &   \\
    \midrule
    IEMOCAP   & 120 & 31 & 5810 & 1623 & 6  \\
    MELD   & 1153 & 280 & 11098 & 2610 & 7 \\
    \bottomrule[1.2pt]
\end{tabular}
}
\end{table}

\begin{table*}[t]
\centering
\caption{The number of utternces with each emotion label in the IEMOCAP and MELD test dataset.}
\label{tab:2}
\begin{tabular}{c c c c c c c c c c c}
\hline
\multirow{2}{*}{Dataset}&\multicolumn{9}{c}{Emotions} \\
\cline{2-10}

& Neutral & Surprise & Fear & Sad/Sadness & Happy/Joy & Disgust &Angry &Excited & Frustrated \\
\hline
\multicolumn{1}{c}{IEMOCAP} &384	&-	&-	&245	&144	&-	&170 &299 &381	\\ 
\multicolumn{1}{c}{MELD} &1256	&281	&50	&208	&402	&68	&345 &- &-	\\ 
\hline

\end{tabular}
\end{table*}

\textbf{IEMOCAP:} The multimodal ERC dataset. Each conversation in IEMOCAP is from two actors' performances based on the script. There are 7433 utterances and 151 conversations in IEMOCAP. Each utterance in the conversation is labeled with six categories of emotions: \emph{happy, sad, neutral, angry, excited}, and \emph{frustrated}.

\textbf{MELD:} The data were obtained from the TV show \emph{Friends} and included a total of 13708 utterances and 1433 conversations. Unlike the IEMOCAP dyadic dataset, MELD has three or more speakers in a conversation, and each utterance in the conversation is labeled with seven categories of emotions: \emph{neutral, surprise, fear, sadness, joy, disgust}, and \emph{anger}.

\textbf{Evaluation Metrics:} We use the F1-score to evaluate the performance for each emotion class and use the weighted average of accuracy and F1-score(W-F1) to evaluate the overall performance on the two datasets.

\subsection{Baselines}
\begin{itemize}[itemsep=0pt,topsep=0pt,parsep=0pt]
    \item \textbf{BC-LSTM} \cite{bc-lstm} encodes contextual semantic information through a Bi-LSTM network, thus making emotion predictions.
    \item \textbf{ICON} \cite{icon/HazarikaPMCZ18} uses two GRUs to model the speaker's information, additional global GRUs are used to track changes in emotional states throughout the conversation, and a multilayer memory network is used to model global emotional states.
    \item \textbf{DialogueRNN} \cite{dialoguernn/MajumderPHMGC19} models the speaker and sequential information in a conversation through three different GRUs (global GRU, speaker GRU, and emotion GRU).
    \item \textbf{DialogueGCN} \cite{dialoguegcn/GhosalMPCG19} applies GCN to ERC, and the generated features can integrate rich information. RGCN and GCN are both nonspectral domain GCN models for encoding graphs.
    \item \textbf{DialogueXL} \cite{dialogueXl/ShenCQX21} uses the XLNet model for ERC to obtain global contextual information.
    \item \textbf{DialogueCRN} \cite{dialoguecrn/HuWH20} introduces a cognitive phase that extracts and integrates emotional cues from the context retrieved during the perception phase.
    \item \textbf{BiDDIN} \cite{biddin} specializes inter-modal and intra-modal interactive modules for corresponding modalities, as well as models contextual influence with an extra positional attention. It is set under the multimodal scenario and employs separate modality-shared and modality-specific modules.
    \item \textbf{MMGCN} \cite{mmgcn/HuLZJ20} uses GCN networks to obtain contextual information, which can not only make use of multimodal dependencies effectively, but also leverage speaker information.
    \item \textbf{MVN} \cite{mvn/MaWLPZY22} explores the emotion representation of the query utterance from the word- and utterance-level views, which can effectively capture the word-level dependencies among utterances and utterance-level dependencies in the context. 
    \item \textbf{CoG-BART} \cite{cogbart/LiYQ22} uses supervised contrastive learning to better distinguish between similar emotional labels and augments the model's ability to handle context with an auxiliary generation task.
    \item \textbf{MM-DFN} \cite{mmdfn/HuHWJM22} fuses multimodal contextual information by designing a new graph-based dynamic fusion module to fully understand multimodal conversational contexts to recognize emotions in utterances.
    \item \textbf{COGMEN} \cite{cogmen/JoshiBJSM22} is a multimodal context-based graph neural network which using local information ( speaker information) and global information (contextual information) in the conversation.
    \item \textbf{EmoCaps} \cite{emocaps/LiTZZ22}  uses the new structure Emoformer to extract emotion vectors from different modalities and fuse them with sentence vectors into a emotion capsule for emotion prediction of utterance.
\end{itemize}

\subsection{Implementation Details}
Our proposed model is implemented on the Pytorch framework. The hyperparameters are set as follows: the number of layers of the Transformer in MPT is 5 $(l=5)$, where the number of prompt attention heads is 5 $(m=5)$. The coefficient $\lambda_1$ for supervised contrast loss in hybrid contrastive learning is 0.1, and the coefficient $\lambda_2$ for unsupervised contrast loss is 0.05. The dropout in both IEMOCAP and MELD is 0.2. The learning rate in IEMOCAP is 0.0001 and in MELD is 0.0003. Each training and testing procedure was run on a single RTX 3090 GPU, and the reports of our implemented models are based on the average scores of five random runs on the test set.

\subsection{Multimodal Feature Extraction}
In this paper, we use pre-extracted unimodal features following identical extraction procedures as previous methods \cite{dag/ShenWYQ20,cmn/HazarikaPZCMZ18,dialoguecrn/HuWH20}.

\textbf{Text Feature.} To extract better utterance representation with strong representational ability, we use the large general pretrained language model RoBERTa-Large \cite{roberta} for text vector encoding extraction. However, unlike other downstream tasks, we use the transformer structure to encode the utterances without classifying or decoding them. More specifically, for each utterance in the text modal, we precede its token with a special token $[CLS]$ to make it of the form of $\{[CLS],w_{i1},w_{i2},\ldots, w_{in_{i}}\}$. Then we use the pooled embedding result of the last layer of $[CLS]$ as the feature representation of $u_i^{T}$, and finally, we obtain a sentence vector with 1024 dimensions for each utterance.

\textbf{Audio and Visual Feature.} In terms of audio features, OpenSmile \cite{opensmile/EybenWS10} is used with the IS13 comparison profile, which extracted a total of 6373 features for each utterance video, we reduced the dimensionality to 1582 for the IEMOCAP and 300 for the MELD dataset by using a fully connected layer.
The visual facial features were extracted by pretraining on the Facial Expression Recognition Plus (FER+) corpus using DenseNet \cite{densely/huang2017}. This captures changes in the expression of the speakers, which is very important information for ERC. Finally, a 342-dimensional visual feature representation was obtained.

\begin{table*}[t]
\centering
\caption{Results of different models on the IEMOCAP dataset.}
\label{tab:3}
\begin{tabular}{c c c c c c c| c c}
\toprule[1.2pt]
% \hline
\multirow{2}{*}{Model}&\multicolumn{8}{c}{IEMOCAP} \\
\cline{2-9} 

& Happy & Sad & Neutral & Angry & Excited & Frustrated & ACC & W-F1\\
\hline
% \midrule
\multicolumn{1}{c}{BC-LSTM} &43.30	&69.28	&55.84	&61.80	&59.33	&60.20	&-	&59.19 	\\ 
\multicolumn{1}{c}{CMN} &30.38	&62.41	&52.39	&59.83	&60.25	&60.69	&-	&56.13 	\\ 
\multicolumn{1}{c}{DialogueRNN*} &33.18	&78.80	&59.21	&65.28	&71.86	&58.91	&-	&62.75  	\\ 
\multicolumn{1}{c}{DialogueGCN*} &43.57	&80.48	&57.69	&53.95	&72.81	&57.33	&63.22	&62.89 	\\ 
\multicolumn{1}{c}{DialogueXL*} &-	&-	&-	&-	&-	&-	&-	&65.94  	\\ 
\multicolumn{1}{c}{DialogueCRN*} &-	&-	&-	&-	&-	&-	&66.05	&59.19  	\\ 
\multicolumn{1}{c}{BiDDIN} &-	&-	&-	&-	&-	&-	&65.60	&65.30 	\\ 
\multicolumn{1}{c}{MVN*} &55.75	&73.30	&61.88	&65.96	&69.50	&64.21	&65.32	&65.44 	\\ 
\multicolumn{1}{c}{CoG-BART*} &-	&-	&-	&-	&-	&-	&-	&66.18  	\\ 
\multicolumn{1}{c}{MMGCN} &42.34	&78.67	&61.73	&69.00	&74.33	&62.32	&-	&66.22  \\ 
\multicolumn{1}{c}{MM-DFN} &42.22	&78.89	&66.42	&69.77	&75.56	&66.33	&68.21	&68.18	\\ 
\multicolumn{1}{c}{COGMEN} &-	&-	&-	&-	&-	&-	&68.20	&67.60  	\\
\multicolumn{1}{c}{EmoCaps} &\textbf{71.91}	&85.06	&64.48	&68.99	&\textbf{78.41}	&66.76	&-	&71.77	\\
\hline
\multicolumn{1}{c}{Ours(MPT-HCL)} &58.13	&\textbf{85.97}	&\textbf{66.75}	&\textbf{69.96}	&74.06	&\textbf{69.06}	&\textbf{72.83}	&\textbf{72.51}	\\ 
\bottomrule[1.2pt]
% \hline
\end{tabular}
\end{table*}

\begin{table*}[t]
\centering
\caption{Results of different models on the MELD dataset.}
\label{tab:4}
\begin{tabular}{c c c c c c c c| c c}
\toprule[1.2pt]
% \hline
\multirow{2}{*}{Model}&\multicolumn{9}{c}{MELD} \\
\cline{2-10}

& Neutral & Surprise & Fear & Sadness & Joy & Disgust &Angry &ACC & W-F1 \\
\hline
\multicolumn{1}{c}{BC-LSTM} &73.80	&47.70	&5.40	&25.10	&51.30	&5.20	&38.40 &59.62 &57.29	\\ 

\multicolumn{1}{c}{DialogueRNN*} &73.50	&49.40	&1.20	&23.80	&50.70	&1.70	&41.50 &60.31 &57.66	\\ 
\multicolumn{1}{c}{DialogueGCN*} &-	&-	&-	&-	&-	&-	&- &58.62 &56.36 \\ 
\multicolumn{1}{c}{DialogueXL*} &-	&-	&-	&-	&-	&-	&- &-	&62.41	\\ 
\multicolumn{1}{c}{DialogueCRN*} &-	&-	&-	&-	&-	&-	&- &61.11 &58.67 \\ 
\multicolumn{1}{c}{MVN*} &76.65	&53.18	&11.70	&21.82	&53.62	&21.86 &42.55	&61.29	&59.03	\\ 
\multicolumn{1}{c}{CoG-BART*} &-	&-	&-	&-	&-	&-	&- &-	&64.81	\\ 
\multicolumn{1}{c}{MMGCN} &-	&-	&-	&-	&-	&-	&- &60.42 &58.31	\\ 
\multicolumn{1}{c}{MM-DFN} &77.76	&50.69	&-	&22.93	&54.78 &-	&47.82	&62.49	&59.46	\\ 
\multicolumn{1}{c}{EmoCaps} &77.12	&\textbf{63.19}	&3.03	&42.52	&57.50 &7.69	&57.54	&-	&64.00	\\
\hline
\multicolumn{1}{c}{Ours(MPT-HCL)} &\textbf{77.82}	&58.26	&\textbf{21.52}	&\textbf{45.15}	&\textbf{60.18}	&\textbf{30.36}	&\textbf{59.25}	&\textbf{65.86} &\textbf{65.02}	\\ 
% \hline
\bottomrule[1.2pt]
\end{tabular}
\end{table*}

\section{Results and Analysis}

\subsection{Main Result}
Tables~\ref{tab:3} and Tables~\ref{tab:4} show the experimental results of our proposed MPT-HCL model and the baselines on the IEMOCAP and MELD datasets. Models marked with "*" only use the text modality, and "-" indicates that the results were not reported. The best results are highlighted in bold. All other results are reported in their respective papers. On the one hand, compared to existing methods, our model achieved better utterance representations through pre-trained language models in sentence encoding. On the other hand, as shown in Tables~\ref{tab:3} and Tables~\ref{tab:4}, we found that: (1) Our proposed MPT-HCL outperforms all baseline models in both evaluation metrics, demonstrating the effectiveness of our model in multimodal ERC. (2) MPT-HCL achieves the best results in almost all emotion categories when compared to EmoCaps. Taking the MELD dataset as an example, our approach outperforms EmoCaps by a large margin in predicting labels with few samples, such as Fear and Disgust. This demonstrates the effectiveness of our multimodal fusion method and the specifically designed hybrid contrastive learning strategy for handling labels with few samples.  (3) The overall effect of MPT-HCL is better than MM-DFN, which is a recent model using speaker information.This highlights the effectiveness of utilizing dialogue information in our approach. We consider the reasons for emotional changes in human communication in real life, which are often influenced either by the words spoken by others or by one's own mood. Therefore, we extract contextual cues from both the speaker's own utterances(Speaker-aware) and the interaction between speakers(Context-aware), which leads to superior results compared to MM-DFN.

% Furthermore, in predicting similar labels, such as Disgust and Angry, our approach significantly outperforms MM-DFN, which performs label merging. These results highlight the effectiveness of our model in addressing the challenges faced by previous studies.
\begin{table}[t]
\centering
\caption{Performance of MPT-HCL under different multimodal settings.}
\label{tab:5}
\begin{tabular}{c c c c c }
\hline
\multirow{2}{*}{}&\multicolumn{2}{c}{IEMOCAP}&\multicolumn{2}{c}{MELD} \\
\cline{2-5}

& ACC & W-F1 & ACC & W-F1 \\
\hline
\multicolumn{1}{c}{Only T} &68.92	&68.35	&61.89	&61.63	\\ 
\multicolumn{1}{c}{Only A} &46.58	&47.26	&44.12	&40.15	\\
\multicolumn{1}{c}{Only V} &39.28	&39.75	&36.25	&35.69	\\
\multicolumn{1}{c}{T+A} &70.09	&69.81	&62.84	&61.12	\\
\multicolumn{1}{c}{T+V} &69.79	&68.31	&61.11	&60.03	\\
\multicolumn{1}{c}{T+A+V} &71.02	&70.22	&62.55	&62.13	\\
\multicolumn{1}{c}{Ours} &\textbf{72.83} &\textbf{72.51} &\textbf{65.86} &\textbf{65.02}\\
\hline
\end{tabular}
\end{table}

\subsection{Ablation Study}
\subsubsection{Various Modalities}
Table~\ref{tab:5} shows the performance of our model on the MELD and IEMOCAP datasets under different modal combinations. We can observe that: (1) The performance of multimodal inputs is better than that of unimodal inputs. Furthermore, among the three modalities of text, audio, and visual, just as we intended to use text as the main modality and other auxiliary modality features as prompt, the text modality performs better than the other two modalities. (2) After adding the text modality, the combination of visual and audio shows a significant improvement in performance. This suggests that the text modality plays an important role in ERC, while audio and visual serve as auxiliary features to improve the accuracy of the model's recognition. This is consistent with our goal of filtering the audio and visual modalities to extract more effective features. (3) When comparing the normal modal combination (i.e., T+A+V) with our feature-filtered modal combination, we found that the latter is more effective for modality interaction. This is because the visual and audio modals were originally used as auxiliary modalities. By further filtering their features and retaining their useful high-level features as prompt information for the text modality, we obtained a more effective multimodal fusion feature representation.

\begin{table}[t]
\centering
\caption{Ablation studies on various modules.}
\label{tab:6}
\begin{tabular}{c c c c c }
\hline
\multirow{2}{*}{}&\multicolumn{2}{c}{IEMOCAP}&\multicolumn{2}{c}{MELD} \\
\cline{2-5}

& ACC & W-F1 & ACC & W-F1 \\
\hline
\multicolumn{1}{c}{Full A} &71.43	&71.08	&64.44	&64.17	\\ 
\multicolumn{1}{c}{Full V} &71.13	&69.89	&64.11	&64.02	\\
\hline
\multicolumn{1}{c}{w/o MPT} &45.57	&45.98	&43.91	&42.68	\\
\multicolumn{1}{c}{w/o UCL} &71.27	&71.11	&63.91	&63.62	\\
\multicolumn{1}{c}{w/o UCL} &71.34	&71.23	&64.41	&63.25	\\
\multicolumn{1}{c}{w/o Sa/Ca-RGCN} &64.10	&63.54	&61.30	&62.22	\\
\multicolumn{1}{c}{Ours} &\textbf{72.83} &\textbf{72.51} &\textbf{65.86} &\textbf{65.02}	\\

\hline
\end{tabular}
\end{table}

\subsubsection{Module Analysis}
To study the contribution of each module in MPT-HCL, we performed an ablation study on both datasets. We consider the following settings:
\begin{itemize}
% [itemsep=0pt,topsep=0pt,parsep=0pt]
    \item \textbf{Full A:} We do not use MFF to filter audio feature.
    \item \textbf{Full V:} We do not use MFF to filter visual feature.
    \item \textbf{w/o MPT:} We remove the multimodal prompt transformer module.
    \item \textbf{w/o UCL:} We remove the unsupervised contrastive learning module.
    \item \textbf{w/o SCL:} We remove the supervised contrastive learning module.
    \item \textbf{w/o Sa/Ca-RGCN:} We remove the module used to extract emotional cues from the text modal.
\end{itemize}
Table~\ref{tab:6} shows the results of our ablation experiments, from which we can conclude that: (1) With Full A/V, we can see that unfiltered modality features introduce a lot of noise when exchanging information across modalities, and the visual modality has more noise than the audio modality due to the complex scenes in the data. (2) We can observe that not using MPT yields poor performance. This can be attributed to the lack of information interaction between modalities and the inability of simple concatenation to leverage complementary effects. It introduces significant noise and leads to a confused feature representation, resulting in subpar performance.  (3) Our proposed HCL is very effective as removing any contrastive loss leads to a decrease in model performance. This module performs better on the MELD dataset, as this dataset has more minority class labels. (4) After removing the Sa/Ca-RGCN emotion cue extraction module, a significant performance drop was observed on both datasets. This is because both datasets involve conversation between two or more speakers, and capturing emotional cues from speakers and contextual information in conversation is crucial. However, although we constructed a speaker-aware relational graph and utilized the relationships between speakers, we did not fully explore the independent information of each speaker, especially in the case of MELD, which consists entirely of multi-party dialogues. Therefore, the performance of this module on the MELD dataset was not as significant as that on the IEMOCAP dataset.

\subsection{The Potency of Hybrid Contrastive Learning}
\begin{figure}[htbp]
\centering
\subfigure[Before HCL]{
\begin{minipage}[t]{0.50\linewidth}
\centering
\includegraphics[width=1.7in]{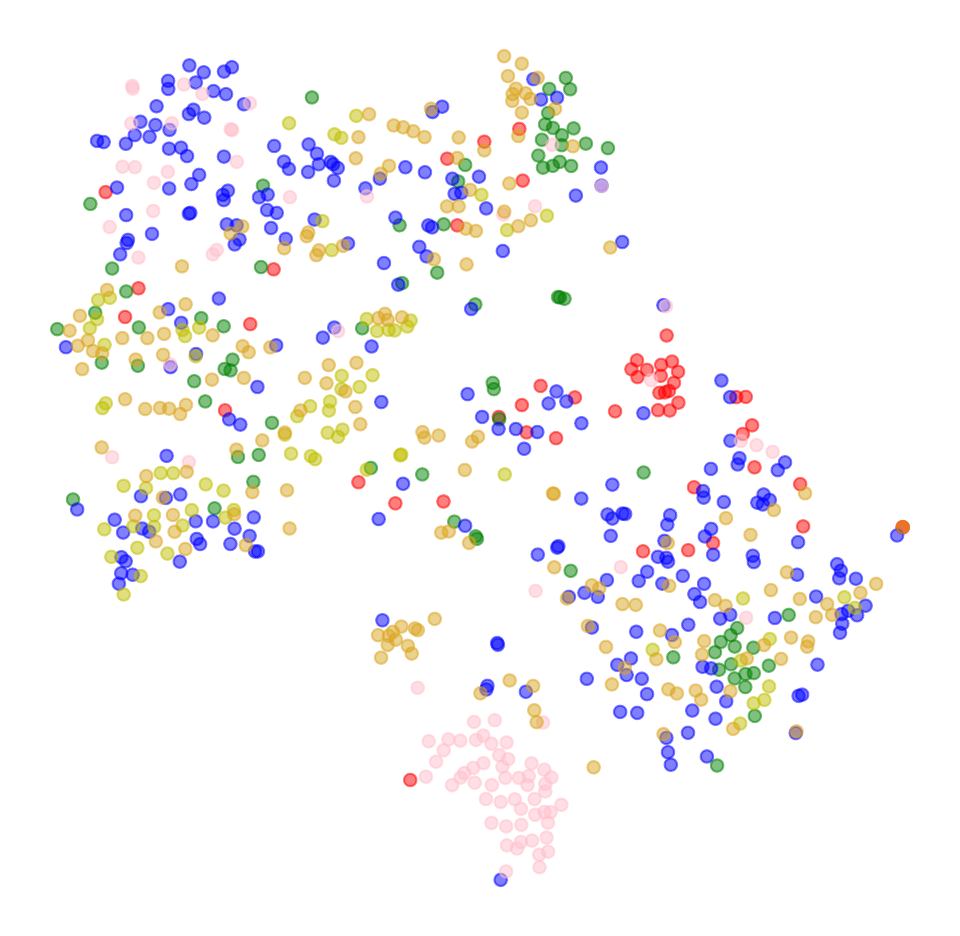}
\end{minipage}
}%
\subfigure[After HCL]{
\begin{minipage}[t]{0.50\linewidth}
\centering
\includegraphics[width=1.7in]{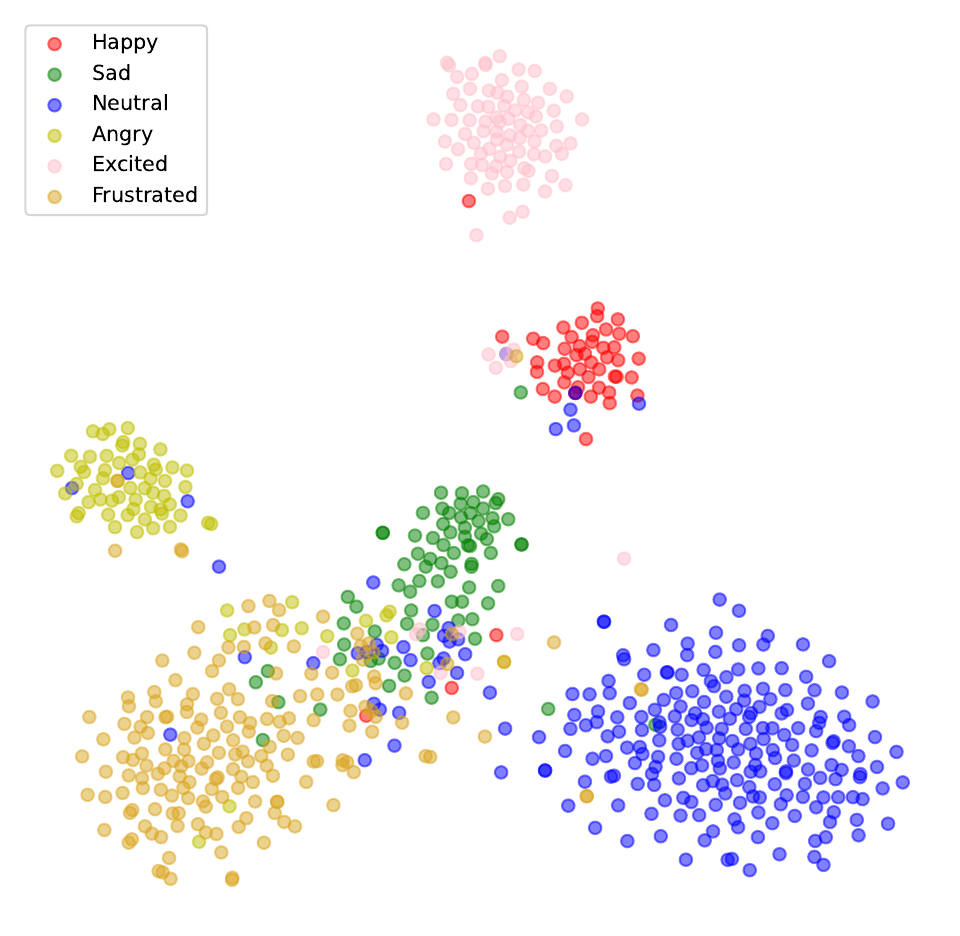}
\end{minipage}
}%
\centering
\caption{The t-SNE visualization results of the model output when HCL is use or not.}
\label{fig:3}
\end{figure}

In order to conduct a qualitative analysis of the hybrid contrastive learning(HCL), we used t-SNE \cite{tsne/hinton2002} to visualize the initial distribution of some data and the hidden layer status of the model after using HCL. As shown in Figure~\ref{fig:3}, when HCL loss is not used, the samples between different labels are randomly scattered, and some samples with similar emotions also overlap, which increases the difficulty of the model learning decision boundaries and leads to large errors in predicting similar labels. With the use of HCL, it can be clearly seen that the coupling degree between different categories gradually increases, resulting in a significant category aggregation effect. This shows that the hybrid contrastive learning strategy we designed plays an important role in sample classification. It is worth noting that although our hybrid contrastive learning has already achieved good category division of samples, we can still see obvious errors in some samples. We analyzed the reason to be that the number of this sample compared to the other labeled samples in the entire batch was the highest, which led to the problem of predicting other labels as this category, and this is a problem that we need to further address.

\begin{figure*}
  \includegraphics[width=\textwidth]{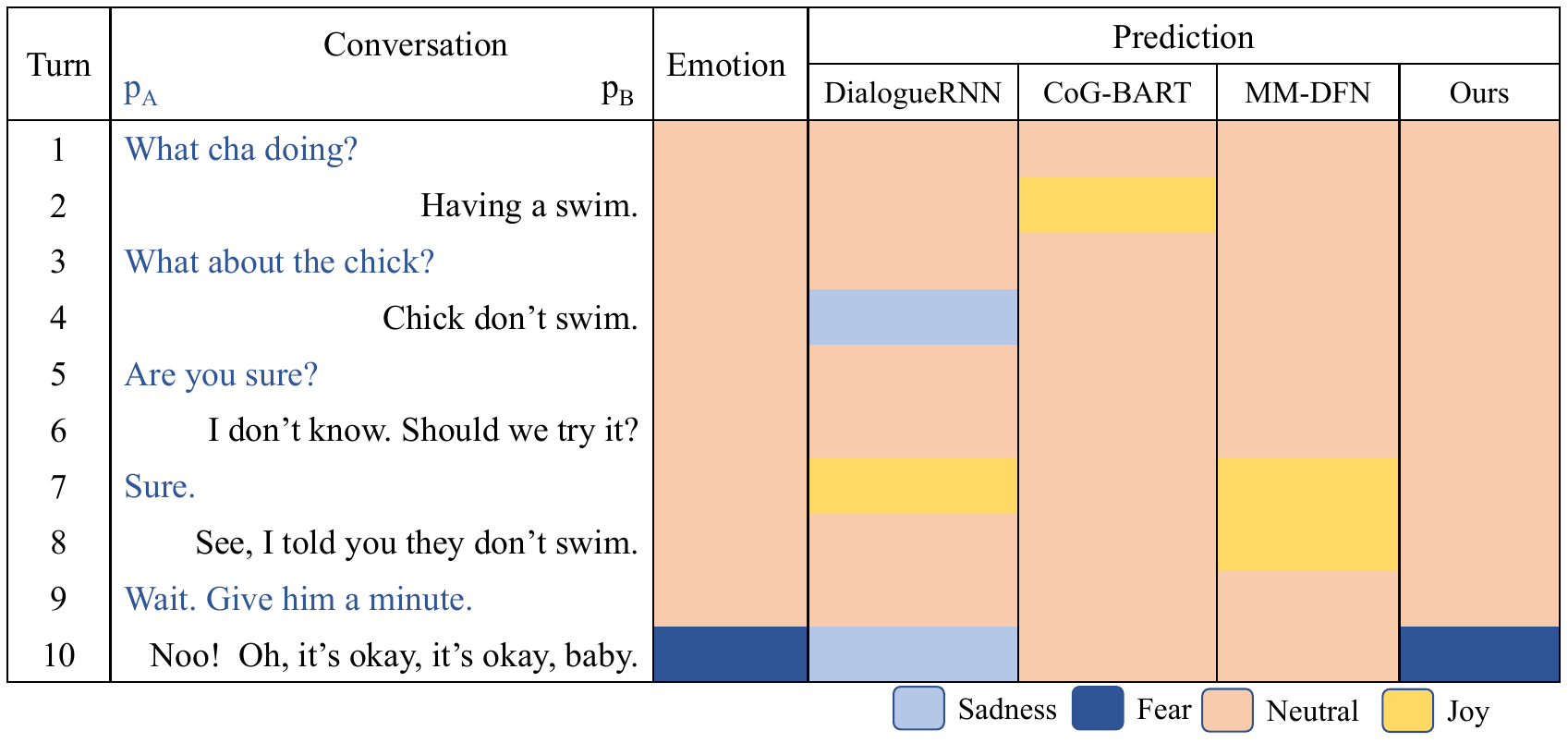}
  \caption{Case study in MELD.}
  \label{fig:4}
\end{figure*}

\subsection{Case Study}
We compare the predictions of different methods for the simultaneous presence of emotion transfer as well as category less labels in utterances. Figure~\ref{fig:4} provides a representative example of the MELD test set. In this example, $P_A$ and $P_B$ try around whether the chick can swim or not, resulting in a final finding that the chick is about to get into danger resulting in an emotion shift from neutral to fear. We observe that our model is more accurate in making emotion predictions compared to the unimodal DialogueRNN and CoG\_BART models for the following reasons: for turn 4, the utterance ends up being incorrectly predicted due to its forward position without much contextual information and without other modal information as an aid. In MM-DFN, for turn 7, the utterance had too little content and the corresponding other modal features could not provide effective help, which led to the model not getting a good fusion representation in the multimodal fusion process of the sentence, resulting in the wrong prediction of the final emotion. Our model first enhances the extracted conversational context with two levels of emotional cue extraction to avoid the problem of capturing insufficient feature content, and also adds less sample labels to the model when predicting the emotion after obtaining valid features by using a hybrid contrastive learning method. Therefore, our model achieves good results in the whole prediction process in both emotion transfer and few-sample prediction.

\section{Conclusion}
We present a new multimodal prompt transformer with hybrid contrastive learning (MPT-HCL) model for ERC. The RGCN is used to extract emotional clues in the conversation, and different levels of emotional clues are extracted to enhance the text modality. Modal feature filters are designed for the visual and audio modalities to filter features and obtain prompt information for the text modality, through MPT to result in better multimodal fusion feature representation. On this basis, hybrid contrastive learning is used to optimize the fusion feature representation and to explore the information of few labels in the samples, thereby improving the prediction accuracy of few sample labels. In the future, we will explore the joint methods for identifying emotional causes and emotional recognition to alleviate the problem of error propagation in information.

% \newpage
%%
%% The acknowledgments section is defined using the "acks" environment
%% (and NOT an unnumbered section). This ensures the proper
%% identification of the section in the article metadata, and the
%% consistent spelling of the heading.
\begin{acks}
This work was supported  by National Natural Science Foundation of China (No. 62141201), National Natural Science Foundation of Chongqing (No. CSTB2022NSCQ-MSX1672), Chongqing University of Technology Graduate Education Quality Development Action Plan Funding Results (No. gzlcx20233217). 
\end{acks}
% \newpage
%%
%% The next two lines define the bibliography style to be used, and
%% the bibliography file.
\bibliographystyle{ACM-Reference-Format}
\balance
\bibliography{refer}

%%
%% If your work has an appendix, this is the place to put it.
% \clearpage

\end{document}